\documentclass{article}
\usepackage{ijcai24}

\usepackage{times}
\usepackage{soul}
\usepackage{url}
\usepackage[hidelinks]{hyperref}
\usepackage[utf8]{inputenc}
\usepackage[small]{caption}
\usepackage{graphicx}
\usepackage{amsmath}
\usepackage{amsthm}
\usepackage{booktabs}
\usepackage{algorithm}
\usepackage{algorithmic}
\usepackage{amssymb}
\usepackage[switch]{lineno}
\usepackage{tabulary}
\usepackage{tabularx}
\usepackage{multirow}

\title{AGHINT: Attribute-Guided Representation Learning on Heterogeneous Information Networks with Transformer}

\author{
Jinhui Yuan$^{1}$
\and
Shan Lu$^2$\and
Peibo Duan$^{3}$  \thanks{Corresponding Authors} \And
Jieyue He$^{1}$ \footnotemark[1] \\
\affiliations
$^1$School of Computer Science and Engineering, Southeast University, Nanjing, China\\
$^2$Nanjing FiberHome Tiandi Co., Ltd., Nanjing, China\\
$^3$Department of Data Science and Artificial Intelligent, Faculty of Information Technology, Monash University, Suzhou, China\\
\emails
jinhuiyuan@seu.edu.cn,
bfact.cn@gmail.com,
peibo.duan@monash.edu,
jieyuehe@seu.edu.cn
}

\begin{document}

\maketitle

\begin{abstract}
    Recently, heterogeneous graph neural networks (HGNNs) have achieved impressive success in representation learning by capturing long-range dependencies and heterogeneity at the node level. However, few existing studies have delved into the utilization of node attributes in heterogeneous information networks (HINs). In this paper, we investigate the impact of inter-node attribute disparities on HGNNs performance within the benchmark task, {\it i.e.}, node classification, and empirically find that typical models exhibit significant performance decline when classifying nodes whose attributes markedly differ from their neighbors. To alleviate this issue, we propose a novel Attribute-Guided heterogeneous Information Networks representation learning model with Transformer (AGHINT), which allows a more effective aggregation of neighbor node information under the guidance of attributes. Specifically, AGHINT transcends the constraints of the original graph structure by directly integrating higher-order similar neighbor features into the learning process and modifies the message-passing mechanism between nodes based on their attribute disparities. Extensive experimental results on three real-world heterogeneous graph benchmarks with target node attributes demonstrate that AGHINT outperforms the state-of-the-art.
\end{abstract}

\section{Introduction}

Heterogeneous information networks (HINs) are intricate structures composed of multiple types of nodes and edges, serving as representations of complex relationships between different types of entities \cite{HIN_Survey}. HINs are widely employed across diverse applications, owing to their exceptional capability to represent complex relationships in the real world. Specifically, heterogeneous graph representation learning aims to effectively encode the topology and heterogeneous information in heterogeneous graph nodes into low-dimensional vectors. 

Recently, heterogeneous graph neural networks (HGNNs) in representation learning have attracted increasing attention, and numerous HGNNs have been proposed to capture long-range dependencies and incorporate inherent heterogeneity effectively \cite{HIN_Representation_Survey}. Regarding the architectural design, HGNNs can be classified into two distinct approaches: meta-path free and meta-path based methods. The former addresses long-range dependencies by stacking multiple convolutional layers and considerate heterogeneity through auxiliary modules \cite{HGT,SHGN,HINormer}. Conversely, meta-path based HGNNs employ predefined meta-paths to incorporate higher-order neighbors, capturing heterogeneity by learning relationships among diverse meta-paths \cite{HAN,MAGNN}.

While several homogeneous graph models have utilized node attributes to guide the graph representation learning process and achieved commendable results \cite{TransGNN,FairVGNN}, few studies examine whether node attributes in heterogeneous graph representation learning still significantly affect model performance. In heterogeneous information networks, nodes of various types often possess attributes with distinct characteristics. As a result, the attribute processing methods applicable in homogeneous graphs cannot be straightforwardly applied to HINs and there remains a significant gap in existing studies regarding the exploration of the impact and utilization of node attributes within HGNNs.

In this work, we conduct an analysis of typical models on target nodes in HINs, focusing on how varying levels of inter-node attribute disparities impact their effectiveness. We utilize a benchmark task, specifically semi-supervised node classification on HINs, as our evaluative framework. Our empirical findings reveal that existing HGNN models struggle to classify nodes effectively when their attributes significantly diverge from those of their neighboring nodes within the receptive field. Motivated by the limitation, we proposed a novel heterogeneous graph model Attribute-Guided Heterogeneous Information Networks representation learning model with Transformer (AGHINT), which is composed of an attribute-guided transformer (AGT) module to enhance node representations with attribute-similar node information and an attribute-guided message weighting (AGM) module to optimize the message-passing mechanism based on target node attributes. In particular, AGHINT harnesses attribute disparities between target nodes to identify their most and least similar neighbors of the same type, along with neighbors of different types via the shortest paths. All target nodes with their identified neighbors will be processed through the above two modules in an end-to-end manner, allowing our model to concentrate on neighbors with similar attributes during the learning process.

Our contribution can be summarized as follows:

\begin{itemize}
\item We introduce an attribute-guided similar node completion module, employing a transformer architecture to enrich the model with information from attribute-similar nodes, and an attribute-guided message weighting module to optimize the interaction process between nodes with differing attributes.
\item We propose a novel heterogeneous graph representation learning model AGHINT built upon the above two modules, which allows us to effectively capture long-range dependencies under attribute guidance beyond the original topology and reduce the impact of dissimilar nodes during the learning process.
\item We conduct extensive experiments on three benchmark datasets with node attributes to evaluate the performance of AGHINT. The results demonstrate the superiority of our model by comparing it with state-of-the-art baselines for heterogeneous graph semi-supervised node classification tasks.
\end{itemize}

\section{Related Work}
\subsubsection{Heterogeneous Graphs Neural Networks}
Traditional graph neural networks (GNNs) have achieved significant advancements in representation learning on homogeneous graphs, which assume node/edge types are singular. However, real-world graphs are often heterogeneous, encompassing multiple types of nodes and edges. Numerous HGNNs have been recently proposed for application on HINs to capture the rich semantic information and complex inter-type relationships. HGNNs can be classified into two categories: meta-path free HGNNs and meta-path based HGNNs. Generally, meta-path based HGNNs employ a hierarchical aggregation process. They first aggregate embeddings of neighbor nodes specified by metapaths to capture certain semantic meanings, and then aggregate these various semantic representations of the target node to produce the final representation, including HAN \cite{HAN}, MAGNN \cite{MAGNN}, GTN \cite{GTN}, etc. This category of HGNNs relies on manually predefined metapaths, posing challenges in generalizability within complex networks.
Conversely, meta-path free HGNNs employ type-aware modules directly within the original HINs to capture complex semantic and structural information. RGCN \cite{RGCN} addresses heterogeneity through relation-specific weights, while HGT \cite{HGT} introduces heterogeneous attention scores in the message-passing mechanism for different node/edge types. Similarly, models like SHGN \cite{SHGN}, HINormer \cite{HINormer} integrate modules specifically designed to capture heterogeneous information. However, these modules commonly overlook the utilization of node attributes.
\subsubsection{Graph Transformers}
Existing studies have increasingly identified key constraints in message-passing-based GNN models, including over-smoothing \cite{Oversmooth} and over-squashing \cite{OverSquash} issues.  The fully connected global attention mechanism of Transformer \cite{Transformer} has demonstrated remarkable efficacy across various graph representation learning tasks. Consequently, integrating the Transformer to enhance GNNs beyond their inherent receptive field limitations is garnering increasing attention. Certain Graph Transformer models, such as GraphTrans [49] and GraphiT [29], strategically combine with GNNs to effectively capture local structural information. Several approaches suggest incorporating graph positional and structural encoding into Transformer \cite{Graphormer} and some of them directly utilize GNNs as structural encoders \cite{SAT}. Recent studies introduce an attribute-based sampling module in their fusion of Transformer and GNNs \cite{TransGNN}, primarily confined to homogeneous graph tasks. Concurrently, some other research targets Graph Transformer modules for heterogeneous graph tasks, developing specialized modules to capture heterogeneity \cite{HINormer}. However, these approaches neglect the significance of node attributes, simply feeding nearest neighbors or all nodes into the Transformer within the original receptive field constraints inherent in GNNs.

\section{Preliminary}
\subsection{Heterogeneous Information Network}
Heterogeneous information networks (HINs) can be defined as graphs \( G = \{V, E, U, R, \mathbf{X}\} \), where \( V \), \( E \), \( U \) and \( R \) correspond to sets of nodes, edges, node types, and edge types, respectively. Each node \( v \) is assigned a type \( \phi(v) \in U \), and each edge \( e \) with a type \( \psi(e) \in R \), where $\phi$ and $\psi$ are the mapping functions for node and edge types. The set $V^{\phi(v)}$ comprises all nodes of the same type as $v$. A HIN should generally satisfy  \( |U| + |R| > 2 \). The node attribute matrix is represented as $ \mathbf{X} \in {\mathbb{R}}^{|U| \times |V^{\phi(v)}| \times d_{\phi(v)}} $, where $ \mathbf{x}_v \in {\mathbb{R}}^{d_{\phi(v)}} $ signifies the attribute vector of ${\phi(v)}$-type node $ v \in V^{\phi(v)} $, and \( d_{\phi(v)} \) indicates the attribute dimension for node type \( \phi(v) \).

\subsection{Graph Neural Networks}
A graph neural network (GNN) is designed for learning representation vector $\mathbf{h}_v^{(L)} \in {\mathbb{R}}^{d_L}$ of each node after $L$-layer transformations from the graph structure and the input node features $\mathbf{h}_v^{(0)} \in {\mathbb{R}}^{d_0}$. GNNs typically employ a message-passing framework where nodes aggregate feature information from their local neighborhoods, which can be mathematically formalized as
\begin{equation}
 \mathbf{h}_v^{(l)} = \text{Aggr}^{(l)} \left( \left\{ \mathbf{h}_u^{(l-1)} : u \in \mathcal{N}(v) \right\};{\theta}_g^l \right),
\end{equation}
where $\text{Aggr} \left(\cdot ;{\theta}_g^l \right)$ is aggregation function parameterized by $ {\theta}_g^l$ in layer $l$, and $\mathcal{N}(v)$ is the neighbors set of node $v$. Heterogeneous graph neural networks constitute a specialized subset of graph neural networks that are adept at managing heterogeneity within complex networked data.

\section{Motivation}\label{Motivation_Section}
In this section, we conduct an empirical study to analyze the influence of attribute disparities among nodes on the efficacy of two representative GNNs and transformer in node classification tasks based on the benchmark-setting \cite{SHGN} of {\tt IMDB} and {\tt DBLP} datasets.

To quantitatively assess the disparities in node attributes, the Jaccard distance metric is employed for nodes with discrete attributes, while the cosine similarity measure is utilized for nodes with continuous attributes. Subsequently, for each target node, we systematically sample all neighboring nodes of identical type within its specified $k$-hop range, as determined by the GNN layers. Then we calculate the average attribute differences among these nodes as follows:
\begin{equation}\label{Equation-4}
    Di_{\text{disc}}(\mathbf{x}_i, \mathbf{x}_j) = \frac{|\mathbf{x}_i \cap \mathbf{x}_j|}{|\mathbf{x}_i \cup \mathbf{x}_j|},
    Di_{\text{cont}}(\mathbf{x}_i, \mathbf{x}_j) = 1 - \frac{\mathbf{x}_i \cdot \mathbf{x}_j}{\|\mathbf{x}_i\| \|\mathbf{x}_j\|},
\end{equation}
\begin{equation}
    \bar{\mathbf{D}}_v = \text{Norm}(\frac{1}{|\mathcal{N}_{k}^{\phi(v)}(v)|} \sum_{n \in \mathcal{N}_{k}^{\phi(v)}(v)}Di(\mathbf{x}_v, \mathbf{x}_n)),
\end{equation}

where $\mathbf{x}_i$ denotes the attribute vector of node $i$. The functions $Di_{\text{disc}}(\cdot)$ and $Di_{\text{cont}}(\cdot)$ quantify the attribute disparity value between two nodes for discrete and continuous attributes, respectively. The term $\mathcal{N}_{k}^{\phi(v)}(v)$ specifies the set of $\phi(v)$-type neighbor nodes within a $k$-hop distance from target node $v$, and $\phi(v)$ is the target node type here. $\text{Norm}(\cdot)$ applies Min-Max normalization. The final vector $\bar{\mathbf{D}} \in {\mathbb{R}}^{|V^{\phi(v)}|}$ calculates the average attribute disparities between each target node and its same-type neighbors, with $|V^{\phi(v)}|$ indicating the total count of target nodes.

\begin{figure}
    \centering
    \includegraphics[width=\columnwidth]{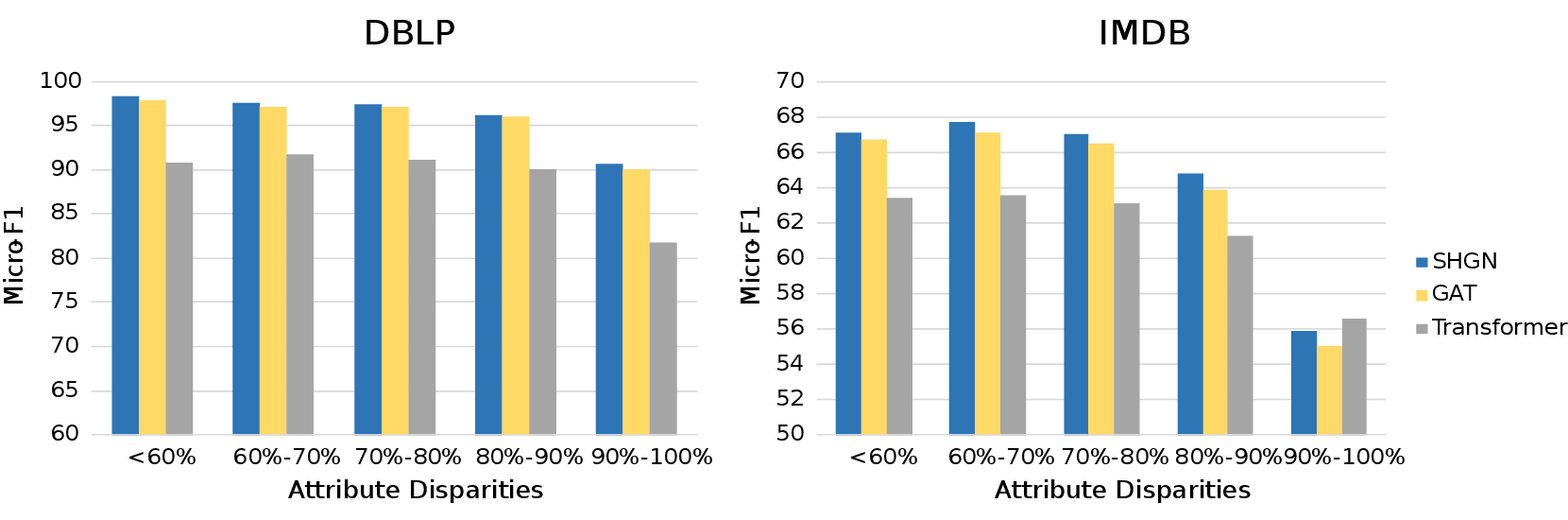}
\caption{Motivation verification on movie and citation datasets. The variation in classification accuracy for target nodes with different neighborhood attribute disparities indicates typical message-passing GNNs show a significant performance decline in classifying target nodes that exhibit significant attribute disparities with their neighboring nodes.}  
\label{fig:Figure2}  
\end{figure}

To examine the effect of neighborhood attribute disparities on the quality of node representations derived by typical models, we distribute each target node into various intervals according to their average attribute disparities $\bar{\mathbf{D}}$ with neighbors. Node classification tasks are then executed for target nodes within each interval. A detailed observation of the performance of GNNs and Transformer on test nodes across varied intervals is illustrated in Figure \ref{fig:Figure2}.

From the result, we observed that message-passing GNNs encounter significant performance declines when classifying target nodes with large attribute disparities from their neighbors. Moreover, neglecting the topological structure and directly inputting sequences of target nodes and their neighbors into a Transformer for updating node representations results in similarly degraded performance, particularly for target nodes associated with sequences that include neighbors with significant attribute differences. We infer that, within HINs, interactions with neighbors having substantial attribute disparities negatively impact the learned node representations, leading to detrimental effects on the model performance.

\section{Methodology}

\begin{figure*}
    \centering
    \includegraphics[width=\textwidth]{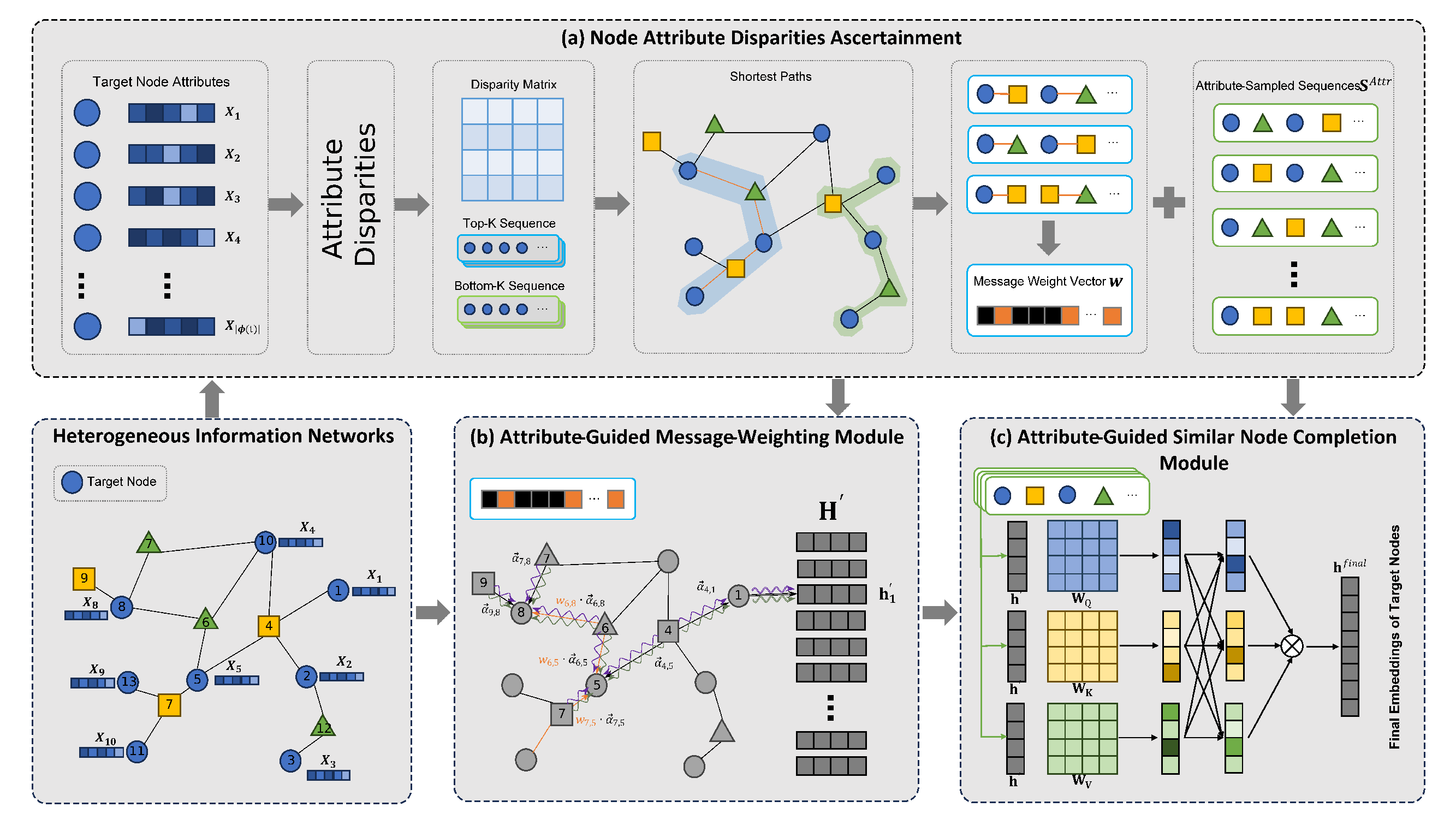}
\caption{The overall architecture of AGHINT.}  
\label{fig:Figure3}  
\end{figure*}

Driven by the above findings, we proposed a novel heterogeneous graph model, AGHINT, featuring an attribute-guided similar node completion module and attribute-guided message weighting module to mitigate the negative impact on learned node representations caused by neighbors with excessively disparate attributes. The comprehensive framework of AGHINT is depicted in Figure \ref{fig:Figure3}. Given a heterogeneous graph $G$ with a full adjacency matrix $A$, we initially compute the attribute between all target nodes to ascertain the inter-node attribute disparities via Equation (\ref{Equation-4}) in Section \ref{Motivation_Section}, which yields a target node attribute disparities matrix. Subsequently, we identify a top-$k$ sequence and a bottom-$k$ node sequence with the most and least attribute disparities for each target node by leveraging this disparities matrix. Then we sample nodes of other types with their interconnecting edges via shortest paths based on the two sequences of each target node. Through the attribute-guided similar node completion module in Fig.2(b) and message weighting module in Fig.2(c), AGHINT can effectively capture long-range dependencies under attribute guidance beyond the confines of the original topology, and adjust the message-passing mechanism based on the node attributes. Finally, node classifications are deduced from the normalized representations engendered by AGHINT, with the model being refined through a supervised classification loss. Subsequent sections will illustrate the details for each part.

\subsection{Node Attribute Disparities Ascertainment}

To ascertain the disparities in node attributes, we initially compare attributes among target nodes following the analysis process outlined in Section \ref{Motivation_Section} and obtain a disparity matrix $\widetilde{\mathbf{D}}$ as

\begin{equation}
    \widetilde{\mathbf{D}}_{ij} = Di(\mathbf{x}_i, \mathbf{x}_j),
\end{equation}
where $\widetilde{\mathbf{D}} \in {\mathbb{R}}^{|V^{\phi(i)}| \times |V^{\phi(i)}|}$ with $\phi(i)$ denoting the target node type and $Di(\cdot)$ is elaborated in Equation (\ref{Equation-4}). It is noteworthy that in heterogeneous graphs, nodes of varying types exhibit attributes with varying dimensions, and it is common to encounter missing attributes in nodes of some non-target type. Consequently, this study focuses on ascertaining the attribute disparities of target nodes.
For each target node $i$, we obtain the top-$k$ nodes and bottom-$k$ nodes from the corresponding $i$-row in the disparity matrix $\widetilde{\mathbf{D}}$ to form the top-$k$ sequences $S^{top}$ and bottom-$k$ sequences $S^{btm}$ for all target nodes.

Through the ascertainment of attribute disparities in target nodes, we have now procured corresponding sequences of same-type neighbors from attribute-similar and dissimilar perspectives.

\subsection{Attribute-Guided Message-Weighting Module}
The objective of this module is to reduce the mutual influence between nodes with widely different attributes. We adjust the message-passing mechanism, which operates on the basis of edges, commences with the construction of a message weight vector 
$\mathbf{w} \in \mathbb{R}^{|E|}$, initialized with ones across all dimensions. Then we determine the shortest paths between each target node and same-type neighbors in the corresponding sequence $S^{top}$ obtained in the previous section. The values in $\mathbf{w}$ corresponding to the edges along these shortest paths are then multiplied by a decay rate ${\alpha}$, as follows:
\begin{equation}
\mathbf{w}_{ij} = \mathbf{w}_{ij} \cdot \alpha, \quad \forall (i, j) \in \text{SP}_G(v_{\text{tgt}}, v_{\text{seq}}),
\end{equation}
where $\text{SP}_G(\cdot)$ is the shortest path function in graph $G$, $v_{\text{tgt}}$ represents the target node, and $v_{\text{seq}}$ denotes the nodes in top-$k$ sequence $S^{top}$. After updating $\mathbf{w}$, we perform a type-specific transformation to get initial features of different node types $\mathbf{H}^{(0)} = \{ \mathbf{h}^{(0)}_0, \mathbf{h}^{(0)}_1, ..., \mathbf{h}^{(0)}_{|V|} \}^{\mathsf{T}} \in \mathbb{R}^{{|V|} \times d_{0} } $ as follows:
\begin{equation}
\mathbf{h}_i^{(0)} = \mathbf{W}_{\phi(i)} \mathbf{x}_i + \mathbf{b}_{\phi(i)},
\end{equation}
where $\mathbf{W}_{\phi(i)} \in {\mathbb{R}}^{d_{\phi(i)} \times d_0}$ represents a learnable parameter matrix, and $\mathbf{b}_{\phi(i)} \in {\mathbb{R}}^{d_0}$ denotes a learnable bias term. Then the node features of varying types are projected into distinct feature spaces. Inspired by \cite{GATv2}, we define the calculation of attention coefficients regarding message weights at each layer as follows:
\begin{equation}
a_{ij} = \frac{\exp\left( \alpha^\mathsf{T} \sigma \left([\mathbf{W} \mathbf{h}_i \Vert \mathbf{W} \mathbf{h}_j] \right) \cdot \mathbf{w}_{ij} \right)}{\sum_{k \in \mathcal{N}(i)} \exp\left( \alpha^\mathsf{T} \sigma \left([\mathbf{W} \mathbf{h}_i \Vert \mathbf{W} \mathbf{h}_k] \right) \cdot \mathbf{w}_{ik} \right)},
\end{equation}
where $\mathbf{w}_{ij}$ is the message weight between node $i$ and $j$, $\mathbf{W}$ represents a learnable matrix, and $\sigma$ is nonlinear activation function. After obtaining the attention coefficients, we compute the attribute-weighted messages with neighborhood features. The aggregation process from messages to updated target node representations in $l$-th layer can be expressed as\:
\begin{equation}
m^{(l)}_{ij} = a^{(l)}_{ij} \mathbf{W}^{(l)} \mathbf{h}^{(l-1)}_j,
\end{equation}
\begin{equation}
\mathbf{h}^{(l)}_i = \sigma \left( \sum_{j \in \mathcal{N}(i)} m^{(l)}_{ij} \right),
\end{equation}
where \( m_{ij}^{(l)} \) represents the message sent from node \( j \) to node \( i \), scaled by the attribute adjustment attention coefficient \( a_{ij}^{(l)} \). \( \mathbf{h}^{(l)}_i \) denotes the updated representation of node \( i \) after summing all incoming messages from the set of neighboring nodes \( \mathcal{N}(i) \). After the encoding of the $L_M$-layer AGM module, the hidden embeddings of all nodes $\mathbf{H}' = \{ \mathbf{h}^{(L_M)}_0, \mathbf{h}^{(L_M)}_1, ...,\mathbf{h}^{(L_M)}_{|V|} \} ^{\mathsf{T}} \in \mathbb{R}^{{|V|} \times d' } $ are obtained, which mitigates the influence of neighboring nodes with significant attribute disparities. \( d' \) signifies the hidden embedding dimension here.

\begin{table*}[ht]
    \renewcommand{\arraystretch}{1.1}
    \centering
    \resizebox{1.0\textwidth}{!}{
    \begin{tabulary}{\textwidth}{l|ccccccc}
        \toprule
        Datasets& Nodes& Edges& Node Types& Edge Types& Target&Classes  &Target Attributes\\
        \midrule
        DBLP& 26,128& 239,566& 4& 6& author& 4 &334\\
        IMDB& 21,420& 86,642& 4& 6& movie& 5 &3489\\
        ACM& 10,942& 547,872& 4& 8& paper& 3 &1902\\
        \bottomrule
    \end{tabulary}}
    \caption{Statistics of the datasets}
    \label{Datasets}
\end{table*}

\subsection{Attribute-Guided Similar Node Completion Module}
For target nodes that exhibit significant attribute disparities with their neighboring nodes, AGHINT employs a Transformer-based similar node completion module to tackle the challenge of a scarcity of similarly attributed nodes in their neighborhood.

To acquire attribute-similar nodes of the same type beyond the immediate neighborhood as well as related nodes of differing types where attribute comparison is not feasible, for each target node, we sample nodes along the shortest paths between the target node $i$ and others in its corresponding bottom-k sequence $S^{btm}_i$:
\begin{equation}
S^{Attr}_i = \bigcup_{j \in S^{btm}_i} \{ \text{SP}_G (i, j) \},
\end{equation}
where $\text{SP}_G (\cdot)$ identifies the set of nodes located on the shortest path between node \( i \) and \( j \) within graph \( G \).  Following this, we construct attribute-sampled sequences $S^{Attr} = \{ S^{Attr}_0, S^{Attr}_1, ..., S^{Attr}_{|V^{\phi(i)}|} \}$ that effectively capture attribute-similar nodes. To standardize the sequence lengths to facilitate compatibility with Transformer, all input sequences are uniformly truncated to a fixed length $n$.

Consequently, for the sampled sequence associated with a given target node \( i \), denoted by \( S^{Attr}_i = \{ v, v_1, v_2, \dots, v_{n-1} \} \), we input its corresponding representations from $\mathbf{H}'$ to the Transformer. Leveraging the multi-head self-attention (MSA) mechanism  in Transformer, we aggregate information from attribute-similar neighbors in $S^{Attr}_i$ and update the representation of node \( i \), as shown in the following equation:

\begin{equation}
 \mathbf{Q} = {\mathbf{H}'_i}\mathbf{W_Q},  \mathbf{K} = {\mathbf{H}'_i}\mathbf{W_K},  \mathbf{V} = {\mathbf{H}'_i}\mathbf{W_V},
\end{equation}

\begin{equation}
\text{MSA}(\mathbf{H}) = \text{softmax}\left(\frac{\mathbf{QK}^\mathsf{T}}{\sqrt{d_K}}\right)\mathbf{V},
\end{equation}

\begin{equation}
 \mathbf{H}^{(l)} = \text{LN}(\text{MSA}(\mathbf{H}^{(l-1)})+\mathbf{H}^{(l-1)}),
\end{equation}
where $\mathbf{W_Q} \in {\mathbb{R}}^{d' \times d_Q}$, $\mathbf{W_K} \in {\mathbb{R}}^{d' \times d_K}$, $\mathbf{W_V} \in {\mathbb{R}}^{d' \times d_V}$, $\mathbf{H}'_i$ is the hidden embeddings of nodes in $S^{Attr}_i$ and $\text{LN}(\cdot)$ denotes the layer normalization operation. Through an \( L_T \)-layer Transformer, we obtain the final output for all nodes in the attribute-sampled sequence $S^{attr}_i$, denotes as $\mathbf{H}^{(L_T)}_{i}$. The first representation within this output is utilized as the final representation for the target node $i$:

 \begin{equation}
 \mathbf{h}_{i}^{final} = \mathbf{H}^{(L_T)}_{i,1},
\end{equation}

For all target nodes, the nodes within their corresponding attribute-sampled sequences serve as inputs to the Transformer, aiming to update their representations, resulting in ${\mathbf{H}}^{final} \in {\mathbb{R}}^{|V^{\phi(i)}| \times d} $. These resultant target node representations are enriched by integrating additional information from attribute-similar neighbors.

\begin{table*}[htbp]
\renewcommand{\arraystretch}{1.1}
\centering
\resizebox{1.0\textwidth}{!}{
\begin{tabulary}{\textwidth}{l|cccccc}
\toprule
\multirow{2}{*}{\textbf{Methods}}     & \multicolumn{2}{c}{\textbf{DBLP}}            & \multicolumn{2}{c}{\textbf{IMDB}}         & \multicolumn{2}{c}{\textbf{ACM}}  \\
                     & \textbf{Micro-F1}    & \textbf{Macro-F1}   & \textbf{Micro-F1}    & \textbf{Macro-F1}   & \textbf{Micro-F1}    & \textbf{Macro-F1}   \\
\midrule
GCN                  & 91.47 ±0.34           & 90.84 ±0.32         & 64.82 ±0.64          & 57.88 ±1.18         & 92.12 ±0.30          & 92.17 ±0.24          \\
GAT                  & 93.39 ±0.30           & 93.83 ±0.27         & 64.86 ±0.43          & 58.94 ±1.35         & 92.19 ±0.93          & 92.26 ±0.94          \\
Transformer          & 92.45 ±0.63           & 91.75 ±0.72         & 61.91 ±0.21          & 57.89 ±0.50         & 85.06 ±0.31          & 85.04 ±0.36          \\
\midrule
RGCN                 & 92.07 ±0.50           & 91.52 ±0.50         & 62.05 ±0.15          & 58.85 ±0.26         & 91.41 ±0.75          & 91.55 ±0.74          \\
HetGNN               & 92.33 ±0.41           & 91.76 ±0.43         & 51.16 ±0.65          & 48.25 ±0.67         & 86.05 ±0.25          & 85.91 ±0.25          \\
GTN                  & 93.97 ±0.54           & 93.52 ±0.55         & 65.14 ±0.45          & 60.47 ±0.98         & 91.20 ±0.71          & 91.31 ±0.70          \\
HAN                  & 92.05 ±0.62           & 91.67 ±0.49         & 64.63 ±0.58          & 57.74 ±0.96         & 90.79 ±0.43          & 90.89 ±0.43          \\
MAGNN                & 93.76 ±0.45           & 93.28 ±0.51         & 64.67 ±1.67          & 56.49 ±3.20         & 90.77 ±0.65          & 90.88 ±0.64          \\
\midrule
RSHN                 & 93.81 ±0.55           & 93.34 ±0.58         & 64.22 ±1.03          & 59.85 ±3.21         & 90.32 ±1.54          & 90.50 ±1.51          \\
HGT                  & 93.49 ±0.25           & 93.01 ±0.23         & 67.20 ±0.57          & 63.00 ±1.19         & 91.00 ±0.76          & 91.12 ±0.76          \\
SHGN                 & 94.46 ±0.22           & 94.01 ±0.24         & 67.36 ±0.57          & 63.53 ±1.36         & \underline{93.35 ±0.45}  & \underline{93.42 ±0.44}  \\
HINormer             &\underline{94.94 ±0.21} &\underline{94.57 ±0.23} &\underline{67.83 ±0.34} &\underline{64.65 ±0.53} &92.12 ±0.27   & 92.19 ±0.27          \\
\midrule
AGHINT               &\textbf{95.47± 0.13}   &\textbf{95.12 ±0.14} &\textbf{69.30 ±0.23}  &\textbf{66.49 ±0.30} &\textbf{93.98 ±0.34}  &\textbf{94.04 ±0.35}  \\
\bottomrule
\end{tabulary}}
\caption{Comparison on node classification in terms of Micro F1 and Macro F1. The best results are denoted in bold, with the second-best results underlined. The error bars (±) represent the standard deviation of the results over five runs.}
\label{Comparison}
\end{table*}

\subsection{Training Objective}
After obtaining the final representation of target nodes, AGHINT adheres to the standard semi-supervised node classification pipeline, employing a linear layer function parameterized by $\theta_l$ to predict the class distribution. Subsequently, the model utilizes a cross-entropy loss for optimization, as defined by the following equations:
\begin{equation}
\hat{\mathbf{y}}_v = Linear(\mathbf{h}^{final}_v; \theta_l),
\end{equation}
\begin{equation}
\mathcal{L} = \sum_{v \in\mathcal{V}_{train}} CE(y_{v}, \hat{y}_{v}),
\end{equation}
where $\hat{\mathbf{y}}_v$ represents the predicted class distribution for node $v$, and $\mathcal{L}$ is the cross-entropy loss computed over the training node set $\mathcal{V}_{train}$. The function $CE(\cdot)$ represents the cross-entropy loss between the true label $y_{v}$ and the predicted label $\hat{\mathbf{y}}_v$.

\section{Experiment}
In this part, we evaluate the benefits of our proposed AGHINT compared with different models on node classification and further give detailed model analysis from several aspects.

\subsection{Experimental Setups}

\subsubsection{Datasets.}
We test our model AGHINT on three public benchmark datasets with target node attributes, including two academic citation datasets ACM and DBLP, and a movie dataset IMDB. The specifics of these heterogeneous graph datasets are summarized in Table \ref{Datasets}.

\subsubsection{Baselines.}
To comprehensively evaluate the proposed AGHINT against the state-of-the-art approaches, we consider nine HGNN baselines from two main categories: (1) \textit{Meta-path based} HGNNs, encompassing models such as \textbf{RGCN} \cite{RGCN}, \textbf{HetGNN} \cite{HetGNN}, \textbf{HAN} \cite{HAN}, \textbf{GTN} \cite{GTN}, \textbf{MAGNN} \cite{MAGNN}; (2)\textit{Meta-path free} HGNNs, which include \textbf{RSHN} \cite{RSHN}, \textbf{HGT} \cite{HGT}, \textbf{SHGN} \cite{SHGN}, and \textbf{HINormer} \cite{HINormer}. In addition, we incorporate three homogeneous approaches for comparison: \textbf{GCN} \cite{GCN}, \textbf{GAT} \cite{GAT}, and \textbf{Transformer} \cite{Transformer}. To ensure a fair comparison, we configured the hyperparameters for all baselines under experimental settings officially reported by the authors.

\subsubsection{Implementation details.}
We conduct multi-class node classification on DBLP, ACM, while multi-label node classification on IMDB. Following the standardized process pipeline outlined in the heterogeneous graph benchmark (HGB)\footnote{\url{https://www.biendata.xyz/hgb/}}, nodes are randomly divided in a 24:6:70 ratio for training, validation, and testing phases, respectively, and we directly borrow the results reported in HGB for comparison. In cases where HGB does not provide relevant data, we replicate experiments adhering to their original experimental setups.

Micro-F1 and Macro-F1 are both utilized as metrics to assess classification performance. Each experiment is conducted five times to ensure reliability, and the results are presented as averages with standard deviations.

\begin{figure}
    \centering
    \includegraphics[width=\columnwidth]{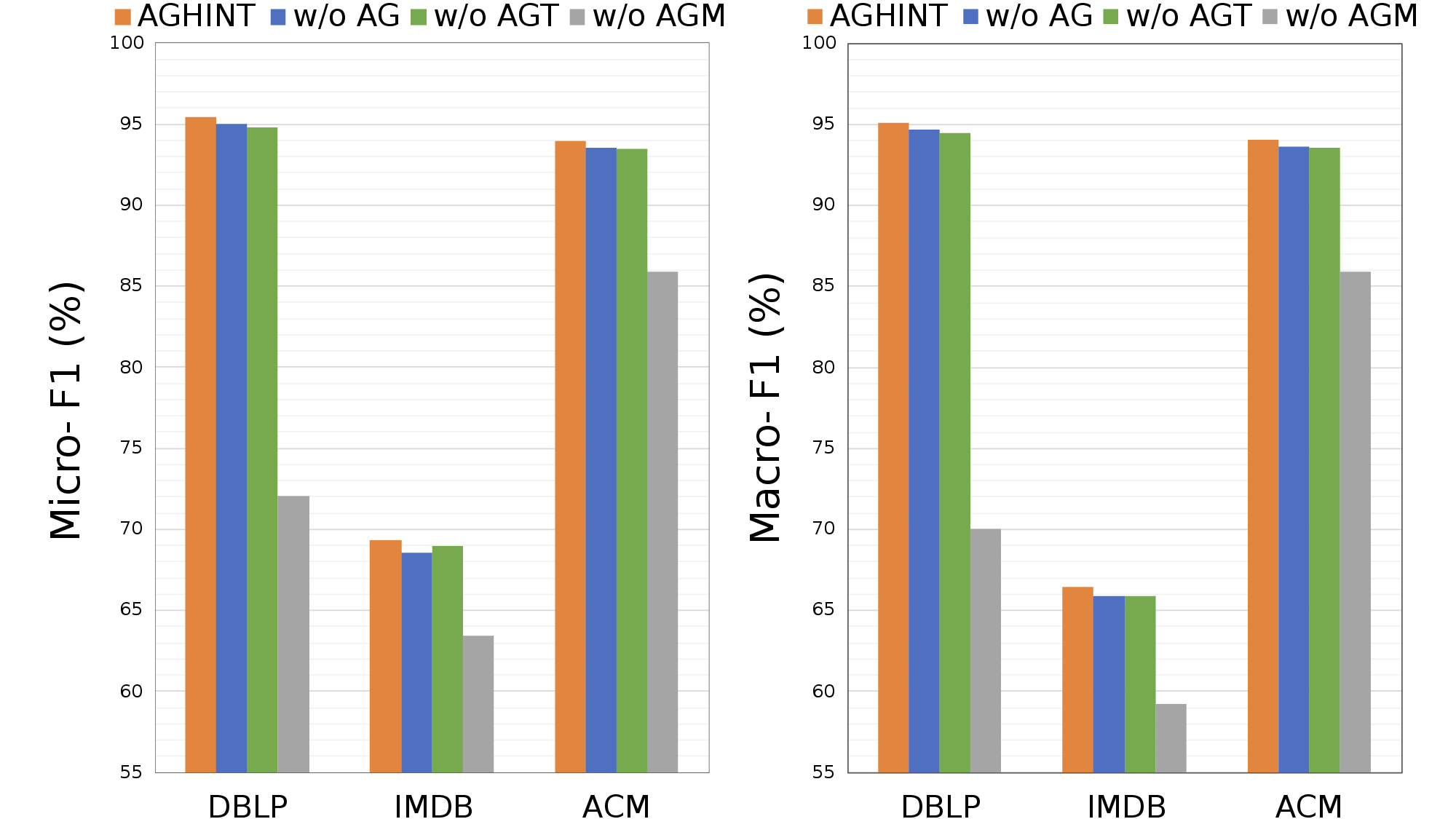}
\caption{Results of different model variants.}  
\label{fig:Ablation}  
\end{figure}

\subsection{Performance Comparison}
In our node classification tasks on three benchmark datasets, we compare the performance of our AGHINT model with other baselines, and as depicted in Table \ref{Comparison}, AGHINT consistently outperforms all competitors.

From the results presented, we make the following observations. Firstly, our model outperforms the vanilla Transformer and GAT, which highlights the effectiveness of our modified attention mechanism and Transformer module, both of which are guided by target node attributes. Secondly, AGHINT exceeds the performance of the graph transformer-based model, HINormer, by a considerable margin of 0.62\% and 1.84\% in Macro-F1 scores on the ACM and IMDB datasets, respectively. This demonstrates the efficacy of node attribute guidance in Transformers for node representation learning. Finally, AGHINT demonstrates a greater performance enhancement on the IMDB dataset, showing a 1.47\% improvement, compared to a 0.53\% improvement on the DBLP dataset in Micro-F1 scores. The more remarkable improvement on IMDB aligns with the more pronounced performance decline observed for IMDB in Section \ref{Motivation_Section} when classifying target nodes that present significant attribute disparities to neighboring nodes. Such a correlation further exemplifies the efficacy of AGHINT in effectively tackling the challenges posed by attribute disparities.

\subsection{Model Analysis}

\begin{figure}
    \centering
    \includegraphics[width=\columnwidth]{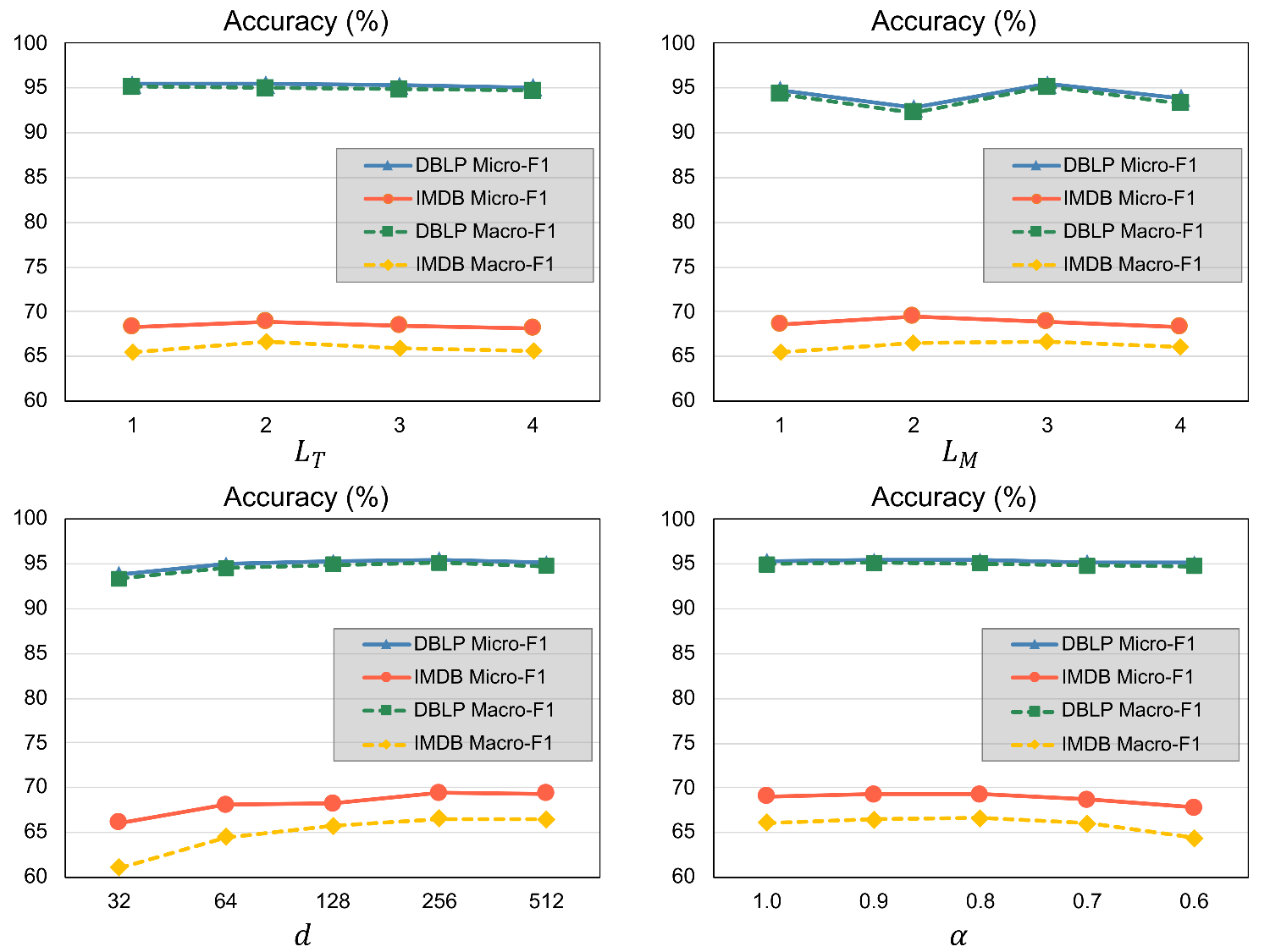}
\caption{Results of different hyperparameter settings.}  
\label{fig:Parameters_Sensitive}  
\end{figure}

\subsubsection{Ablation Study.}

To ascertain the individual contributions of our proposed components, we execute an ablation study across three datasets by comparing with three variants of AGHINT: (1) \textit{w/o.} AGT: we remove the attribute-guided similar node completion module and use the output of attribute-guided message-weighting module as the final node representations; (2) \textit{w/o.} AGM: the attribute-guided message-weighting module is excluded and the projected node embeddings serve as the input representations; (3) \textit{w/o.} AG: We removed the influence of node attributes in all modules, for which we omit the decay rate and use the \textit{D}-hop node context sampling strategy \cite{HINormer} to replace the current sampling strategy based on attribute disparity.

The results of the ablation study are depicted in Fig. \ref{fig:Ablation}. It is evident that the complete AGHINT configuration surpasses the performance of both \textit{w/o.} AGT and \textit{w/o.} AGM variants, confirming the effectiveness of the AGHINT design and the effectiveness of introducing Transformer to supplement node information with similar attributes. Moreover, a notable decline in model performance is observed when the guidance of node attributes is excluded from the message-passing mechanism and the construction of the Transformer input sequence, which shows the efficacy of considering node attribute differences within the model.

\subsubsection{Parameter Sensitivity.}
We study the parameters of AGHINT from four aspects, including the number of module layers $L$, hidden embedding dimension $d$, and decay rate $\alpha$ as shown in Figure \ref{fig:Parameters_Sensitive}.

The results indicate that the performance of AGHINT is notably influenced by the choice of attribute-guided message-weighting module layer $L_M$ and similar node completion module layer $L_T$. The optimal performance is reached approximately at $L_M$ = 2 and $L_M$ = 3 for IMDB and DBLP, respectively. A small number of $L_T$  tends to yield favorable results, which may caused by the small number of nodes in sampled Transformer input sequences. In particular, AGHINT achieves commendable performance with a hidden embedding dimension of 256, and a moderate decay rate $\alpha$ such as [0.9,0.8] appears to benefit the performance.

\subsection{Case Study}

\begin{figure}
    \centering
    \includegraphics[width=\columnwidth]{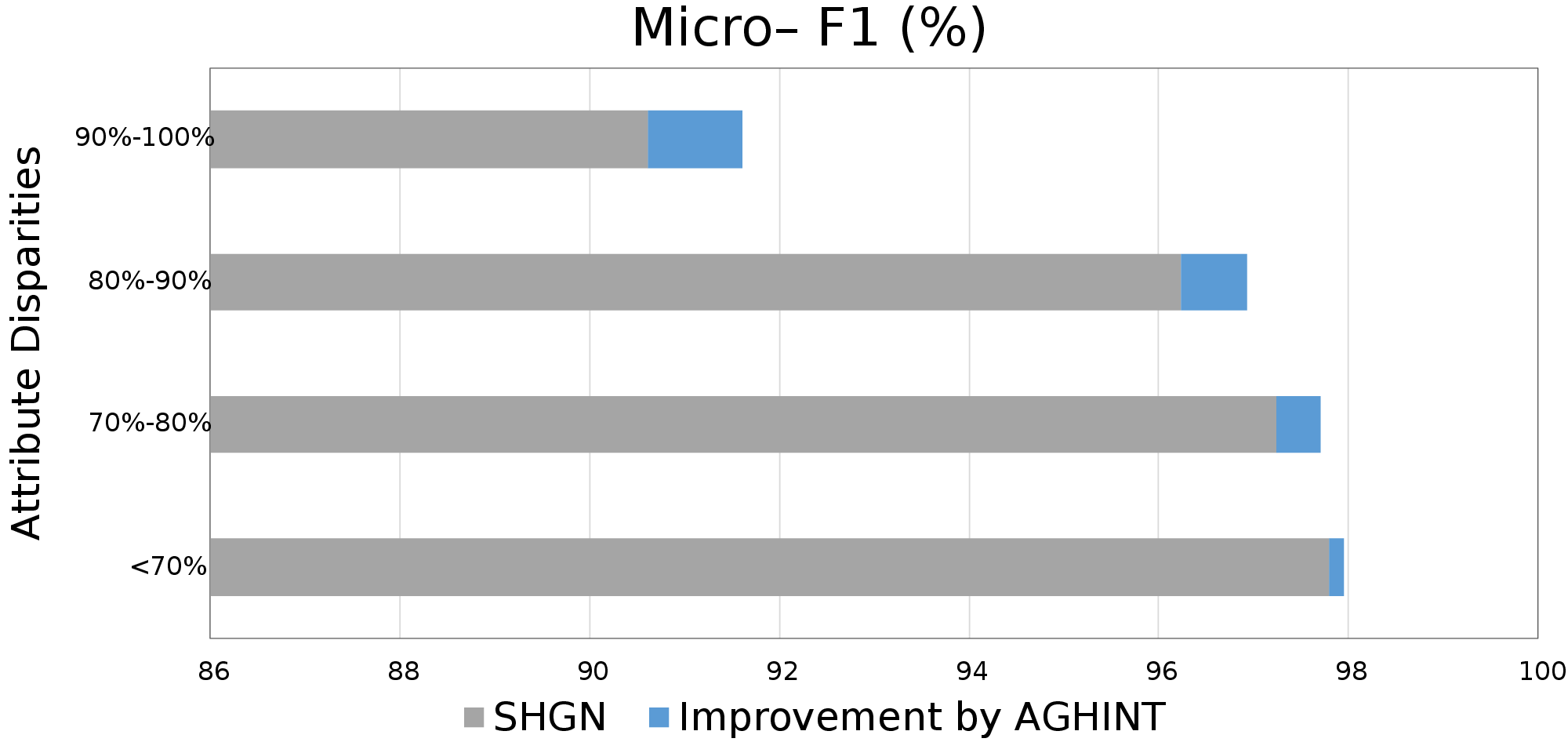}
\caption{Performance improvement by AGHINT on DBLP with varying neighborhood attribute disparities.}  
\label{fig:Case_study}  
\end{figure}

For a more intuitive understanding and comparison, we categorize target nodes into various intervals according to their attribute disparities with neighboring nodes of the same type on DBLP dataset, as detailed in Section \ref{Motivation_Section}. Then we compare the Micro-F1 scores of our proposed AGHINT model against the typical HGNN model, SHGN, within these defined intervals. The results of this comparative analysis are depicted in Figure \ref{fig:Case_study}. The performance improvement of AGHINT over SHGN escalates with the increasing attribute disparities between nodes, achieving its most significant improvement for target nodes with the greatest attribute differences from their neighbors. This indicates that AGHINT, through its attribute-guided modules, effectively assists target nodes in focusing on neighbors with similar attributes, thereby alleviating classification challenges for nodes with significant attribute disparities.

\section{Conclusions}
In this study, we investigate the impact of node attributes on the performance of HGNNs. Our empirical results show typical HGNN models face challenges in effectively classifying nodes with attributes that markedly differ from those of their neighbors. Therefore, we proposed an attribute-guided heterogeneous graph model AGHINT, which contains an optimized message-passing mechanism and a transformed-based module to complement nodes with similar attributes. Extensive experiments conducted on three benchmark datasets with node attributes demonstrate the superiority of the proposed AGHINT over established baselines in semi-supervised node classification tasks and the effectiveness of two submodules. Furthermore, our future work will concentrate on devising methods for constructing higher-quality initial node attributes from raw textual data, building upon the insights gained from this study.

\bibliographystyle{named}

\end{document}